\documentclass[lettersize,journal]{IEEEtran}
\usepackage{amsmath,amsfonts}
\usepackage{algorithmic}
\usepackage{algorithm}
\usepackage{array}
\usepackage[caption=false,font=normalsize,labelfont=sf,textfont=sf]{subfig}
\usepackage{textcomp}
\usepackage{stfloats}
\usepackage{url}
\usepackage{verbatim}
\usepackage{graphicx}
\usepackage{cite}
\usepackage{caption}
\usepackage{csquotes}
\usepackage{tabularx}
\usepackage{multirow}
\usepackage{booktabs}
\usepackage[utf8]{inputenc}
\usepackage[T1]{fontenc}

\begin{document}

\title{How Good is ChatGPT in Giving Adaptive Guidance Using Knowledge Graphs in E-Learning Environments?}

\author{Patrick Ocheja, ~\IEEEmembership{Member, ~IEEE},
Brendan Flanagan, ~\IEEEmembership{Member, ~IEEE}
Yiling Dai,
Hiroaki Ogata, ~\IEEEmembership{Senior Member, ~IEEE}
}



\maketitle

\begin{abstract}
E-learning environments are increasingly harnessing large language models (LLMs) like GPT-3.5 and GPT-4 for tailored educational support. This study introduces an approach that integrates dynamic knowledge graphs with LLMs to offer nuanced student assistance. By evaluating past and ongoing student interactions, the system identifies and appends the most salient learning context to prompts directed at the LLM. Central to this method is the knowledge graph's role in assessing a student's comprehension of topic prerequisites. Depending on the categorized understanding (good, average, or poor), the LLM adjusts its guidance, offering advanced assistance, foundational reviews, or in-depth prerequisite explanations, respectively. Preliminary findings suggest students could benefit from this tiered support, achieving enhanced comprehension and improved task outcomes. However, several issues related to potential errors arising from LLMs were identified, which can potentially mislead students. This highlights the need for human intervention to mitigate these risks. This research aims to advance AI-driven personalized learning while acknowledging the limitations and potential pitfalls, thus guiding future research in technology and data-driven education.
\end{abstract}


\section{Introduction}
\IEEEPARstart{R}{ecent} advances in Artificial Intelligence (AI) have led to transformative approaches in various fields, including education. AI-enabled educational interventions have emerged as powerful tools not only in modeling student behaviors and predicting learning pathways but also in providing dynamic content adaptations \cite{baker2016educational}. Despite these advancements, delivering tailored feedback to students that captures their individual cognitive gaps and learning styles remains a significant challenge \cite{blikstein2016multimodal, ocheja2023visualization}. Large language models (LLMs), such as Generative Pre-trained Transformers (GPTs), have shown broad linguistic comprehension and generation capabilities \cite{radford2019language}, providing potential solutions that align with student-specific misconceptions and learning difficulties. Auto-generated feedback, in particular, stands out as an area where AI and LLMs can truly shine. Shermis and Burstein \cite{shermis2013handbook} investigated automated essay scoring, providing results that suggest AI can offer feedback comparable to human evaluators in specific contexts.

While Intelligent Tutoring Systems (ITSs) have long been explored for personalized learning, they often exhibit limitations in granularity and flexibility. Hwang \cite{hwang2003conceptual} discusses how ITSs typically recommend exercises related to a weak concept but do not provide assistance at the level of addressing specific impasses within an exercise. Additionally, Phobun and Vicheanpanya \cite{phobun2010adaptive} highlight that although some ITSs offer adaptive hints and explanations, these are often predefined and may not be sufficiently flexible to address the changing needs of learners. Early works, such as those by Zhou et al. \cite{zhou1999delivering}, explored delivering hints in a dialogue-based ITS, but the technology used at the time lacked the sophistication of current LLMs, limiting their ability to provide highly personalized and context-aware guidance.

This paper introduces a novel approach that integrates dynamic knowledge graphs with LLMs to offer nuanced student assistance. By evaluating past and ongoing student interactions, the system identifies and appends the most salient learning context to prompts directed at the LLM. Central to this method is the knowledge graph's role in assessing a student's comprehension of topic prerequisites. Depending on the categorized understanding (good, average, or poor), the LLM adjusts its guidance, offering advanced assistance, foundational reviews, or in-depth prerequisite explanations, respectively. This approach aims to overcome the limitations of traditional ITSs by providing more granular and adaptive support tailored to the individual student's needs.

The proposed system leverages concept-map driven approaches \cite{flanagan2019knowledge, ocheja2020prototype} to ascertain and evaluate the importance of prerequisite questions. By incorporating details such as the question, correct and standard solution, the student's impasse, and probable causes retrieved from their current knowledge state (computed from assessment results on prerequisite concepts), we can provide personalized guidance on questions students find problematic. This method not only enhances the precision of feedback but also aligns it more closely with the student's specific learning context. Preliminary findings suggest that students may benefit from this tiered support and could achieve enhanced comprehension and improved task outcomes.

In this paper, we evaluate the proposed methodology using both objective metrics and expert opinions. Specifically, we employ the Recall-Oriented Understudy for Gisting Evaluation (ROUGE) method \cite{lin2004rouge} to compare GPT4-generated feedback across three categories of student performances: low (S1), average (S2), and high (S3), grouped based on their performances on prerequisites. Experts are asked to rate the generated feedback across metrics of correctness, precision, hallucination, and variability. The main research objectives are:
\begin{enumerate}
\item How can we provide personalized guidance to students based on their current impasse and knowledge state?
\item Can LLMs generate correct answers that precisely address the student's learning impasse?
\end{enumerate}

By addressing these questions, this research contributes to the development of more effective AI-driven educational tools, enhancing the personalization and adaptability of e-learning environments.


\subsection{Background}

LLMs are a subset of deep learning models designed to generate human-like responses, summaries, translations, and more by training on vast amounts of text data \cite{li2023ethics}. These models primarily operate based on self-attention mechanisms, a core component of the Transformer architecture introduced by Vaswani et al. \cite{vaswani2017attention}. Self-attention allows LLMs to weigh the significance of different words in a sequence relative to a target word, thereby capturing long-range dependencies and contextual information. This mechanism enables the model to attend to relevant parts of the input text dynamically, improving its ability to understand and generate coherent and contextually appropriate responses.

Mathematically, self-attention is computed as follows: given an input sequence represented by a matrix \(X\), three linear transformations generate the Query \(Q\), Key \(K\), and Value \(V\) matrices. The self-attention output is computed by first calculating the attention scores using the dot product of \(Q\) and \(K\), followed by a softmax operation to obtain the attention weights. These weights are then used to compute a weighted sum of the values in \(V\), as shown in Equation \ref{eq:self-attention}.

\begin{equation}
\text{Attention}(Q, K, V) = \text{softmax}\left(\frac{QK^T}{\sqrt{d_k}}\right)V
\label{eq:self-attention}
\end{equation}

Here, \(d_k\) is the dimension of the Key vectors. This process allows the model to focus on different parts of the input sequence dynamically, enhancing its ability to understand complex language patterns and generate high-quality text \cite{vaswani2017attention}.

Training LLMs involves pre-training and fine-tuning stages. During pre-training, the model learns to predict the next word in a sentence across a diverse and extensive corpus, thereby capturing general language patterns and structures. This phase uses unsupervised learning, where the model optimizes the likelihood of the next word given the preceding words, typically using the cross-entropy loss function. Fine-tuning involves adjusting the pre-trained model on a smaller, task-specific dataset using supervised learning, allowing it to adapt to particular applications or domains. For instance, fine-tuning can be applied to tasks like sentiment analysis, summarization, or machine translation \cite{devlin2018bert}.

A popular example of an LLM is ChatGPT, developed by OpenAI. ChatGPT leverages the GPT architecture, which stands for Generative Pre-trained Transformer \cite{radford2019language}. This model is trained to generate coherent and contextually relevant text based on user input, making it a powerful tool for various applications, including education. We used ChatGPT version 4 (ChatGPT4) in this study as it showed significantly better performance and was deemed a more superior model by OpenAI at the time this study was conducted \cite{plevris2023chatbots}. In educational contexts, ChatGPT can provide personalized assistance, generate explanatory content, and facilitate interactive learning experiences. However, the integration of such technologies in education necessitates a reevaluation of traditional assessment and evaluation methods. Teachers may need to develop new approaches to assess students' understanding and creativity, while students may need to acquire skills in effectively interacting with AI, such as crafting precise prompts to elicit useful responses \cite{dwivedi2023so}.

The advent of LLMs and their specific implementations, such as ChatGPT, has a profound impact on multiple sectors, including education. While there are debates about the benefits and potential drawbacks of using ChatGPT in educational settings, a consensus is emerging on the importance of adapting to this technology. For educators, this involves revising assessment and evaluation strategies to account for the capabilities of AI, while for students, it involves developing new skills to harness the full potential of these tools \cite{stokel2022ai}. Building on these advancements, it is crucial to explore how LLMs can be integrated with existing intelligent tutoring systems (ITSs) and knowledge graph methodologies to provide more granular and adaptive student support. This integration can address the limitations of traditional ITSs and enhance their ability to deliver personalized, context-aware feedback. In the next section, we review related work on ITSs and the use of AI in education, highlighting the unique contributions and gaps that our proposed approach aims to fill.

\begin{figure*}[h]
  \centering
  \includegraphics[width=0.7\linewidth]{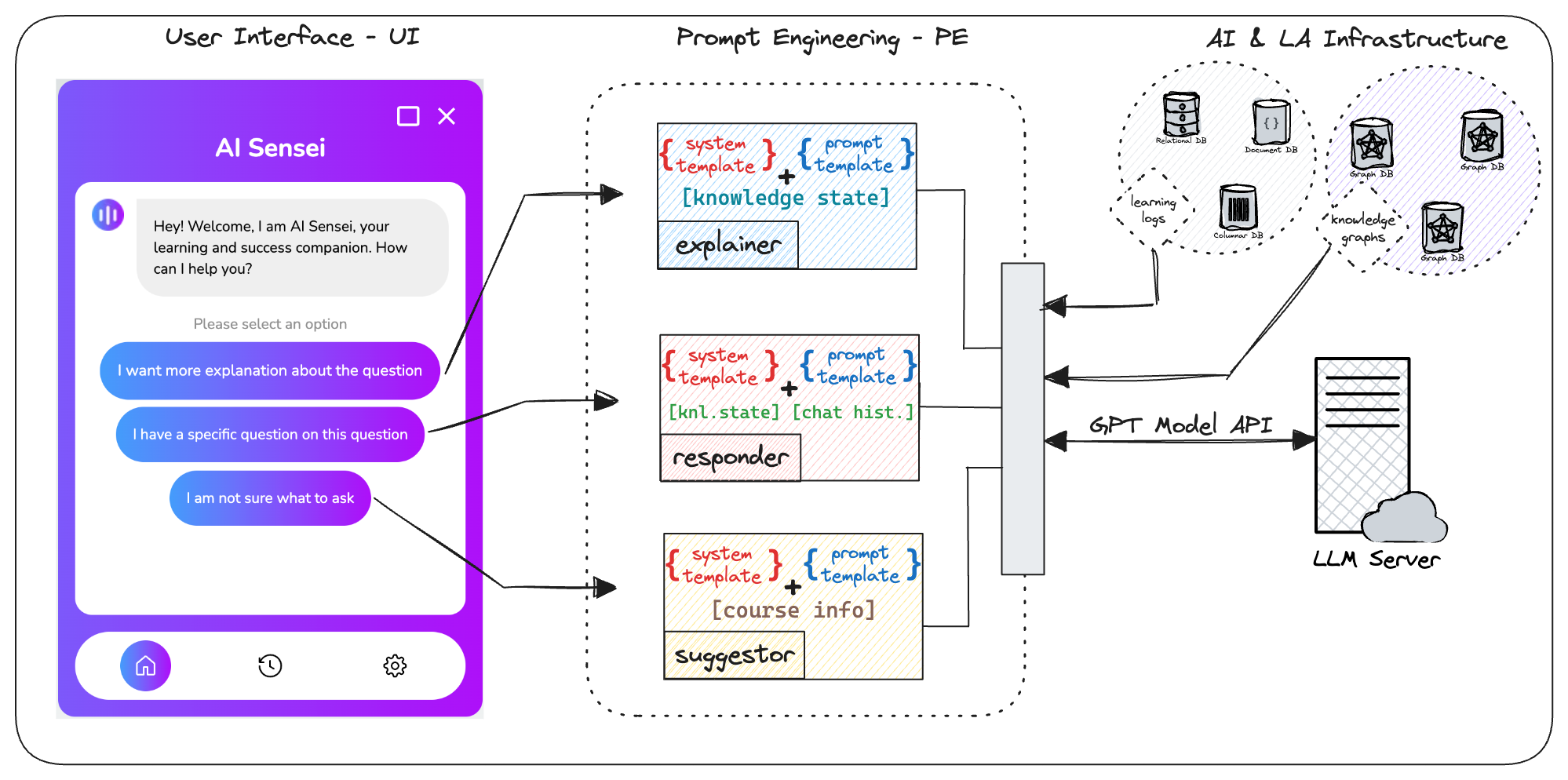}
  \caption{System Architecture.}
  \label{fig:sys_archi}
\end{figure*}

\section{Related Work}

In adaptive learning environments, personalized feedback systems have shown to enhance student engagement and comprehension \cite{siemens2011penetrating}. The integration of AI in education has been extensively researched, with numerous studies highlighting its potential to revolutionize personalized learning experiences. For instance, \cite{holmes2022state, ahmad2020artificial} reviewed the state of AI systems in education, examining their pedagogical and educational assumptions. They revealed a critical insight: many AI tools tend to homogenize student learning experiences rather than offering true personalization. This homogenization occurs because these systems often rely on predefined pathways and solutions, limiting their ability to cater to individual learning needs.

Intelligent Tutoring Systems (ITS) have been a significant area of research, aiming to provide personalized instruction and feedback to students \cite{pai2021application}. Early ITS frameworks primarily focused on rule-based systems that adapted content based on student performance metrics \cite{nwana1990intelligent}. However, these systems struggled with scalability and flexibility \cite{frasson1998designing}. A notable advancement was proposed by Singh, Gulwani and Solar-Lezama \cite{singh2013automated}, who introduced a technique for automated feedback in introductory programming courses. The method in \cite{singh2013automated} utilized a reference implementation to derive corrections for student solutions. Despite its innovation, the technique was limited by its dependency on a correct reference and its inadequacy in addressing large conceptual errors, often resulting in low-level feedback that did not fully support deeper learning processes \cite{baker2016stupid}.

Large Language Models (LLMs) such as GPT-3 and GPT-4 represent a transformative shift in the capability of AI to provide personalized educational support \cite{yan2024practical}. Unlike earlier systems, LLMs can generate nuanced and contextually relevant feedback by understanding the broader context of a student's work. They are capable of recognizing diverse and non-standard solutions due to their training on extensive datasets, offering a more flexible and inclusive approach to feedback \cite{phung2023generating}. For example, in mathematics education, LLMs can provide step-by-step explanations and identify specific misconceptions, thereby tailoring guidance to individual learning paths \cite{yan2024practical}. Recent studies, such as \cite{qadir2023engineering}, have emphasized the potential of LLMs like ChatGPT in delivering personalized feedback. However, these studies also point out the lack of comprehensive experiments and evaluations to substantiate the effectiveness of LLMs in diverse educational settings.

Additionally, the potential of integrating LLMs with knowledge graphs to enhance personalized learning experiences remains underexplored. Knowledge graphs can provide a structured representation of the vast information that LLMs are trained on, facilitating the generation of more precise and relevant feedback \cite{xue2022knowledge,tamavsauskaite2023defining}. They enable the mapping of prerequisite relationships between concepts, allowing LLMs to diagnose knowledge gaps more effectively and deliver targeted support. This synergy between LLMs and knowledge graphs could offer a promising direction for developing advanced ITS that can adapt in real-time to the evolving needs of students.

Despite these technological advancements, there is still a significant gap in the literature regarding the practical application and evaluation of LLMs in educational settings. Most existing studies emphasize the theoretical potential of LLMs, with limited empirical validation. This paper aims to bridge this gap by proposing a comprehensive approach for utilizing LLMs to generate personalized guidance based on students' specific challenges and missing prerequisites. By focusing on concrete experiments and evaluations, we seek to demonstrate how LLMs can be effectively integrated into educational environments to support personalized learning at scale.

To the best of our knowledge, this work is the first to systematically explore and propose a comprehensive framework for the application of LLMs in providing personalized feedback with a focus on mathematics. Our approach leverages the advanced capabilities of LLMs to understand student errors and misconceptions, coupled with the structured insights provided by knowledge graphs, to offer a solution for personalized education. This research aims to bridge the gap between theoretical potential and practical application, providing a foundation for future studies in the field.

\section{Research Methodology}
In this study, we propose a novel method and architecture show in Figure \ref{fig:sys_archi} that combines knowledge graph, learning analytics and LLM-based feedback generation to provide adapted guidance for students based on their specific needs. We called this conversational personalized feedback chatbot AI-sensei. The methodology is structured into the following key phases:

\subsection{Knowledge Graph Creation, Node Identification and Prerequisites}
\begin{figure*}
     \centering
     \includegraphics[width=0.7\linewidth]{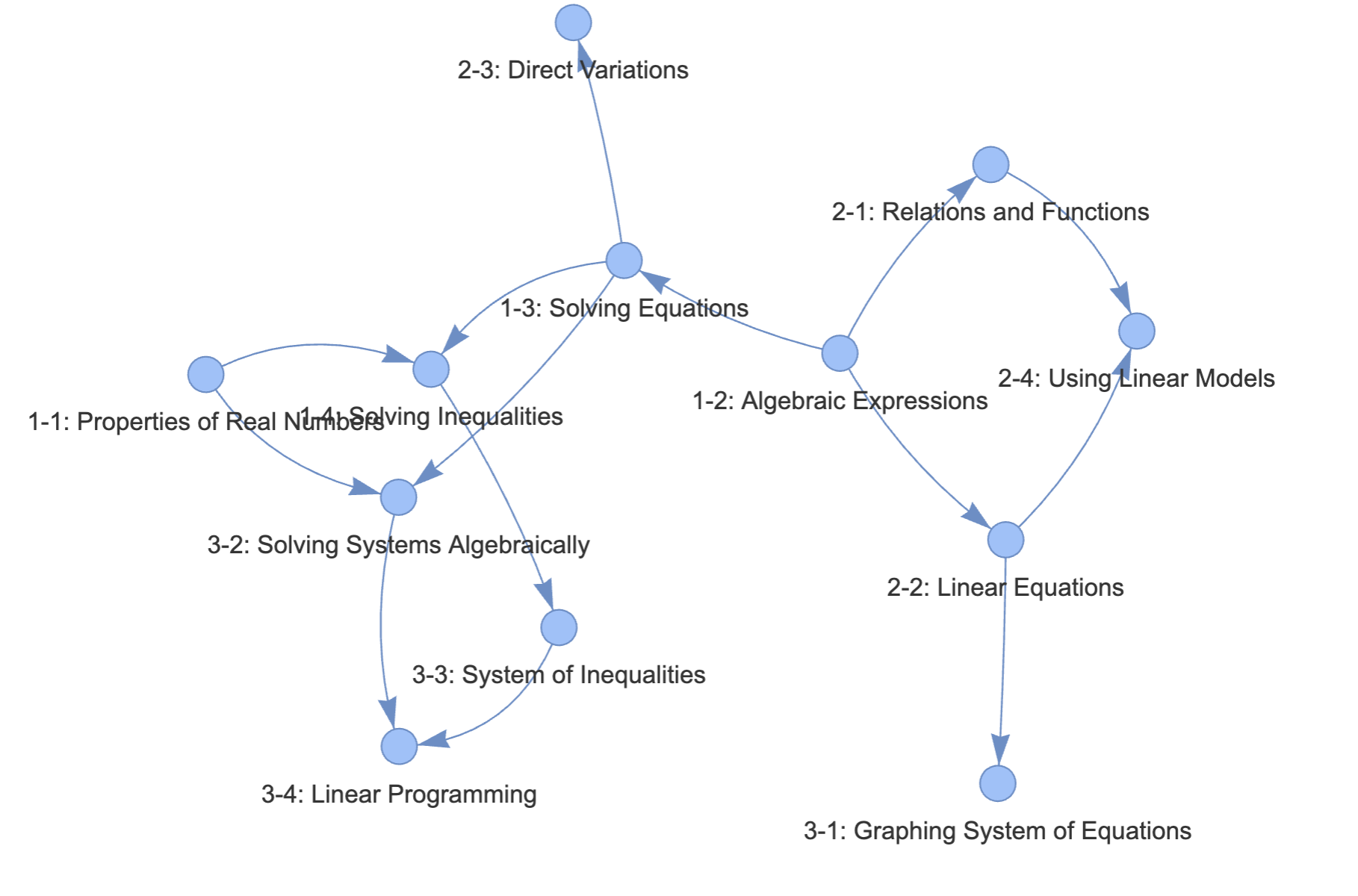}
     \caption{Knowledge graph constructed from the textbook Math Algebra 2 by Prentice Hall}
     \label{fig:kg-math}
\end{figure*}

Our proposed method relies on knowledge graphs \cite{pujara2013knowledge} to determine the relationship and hierarchy of the topics and sub-topics to be learnt by students. In this paper, we use a simplified knowledge graph of Mathematics constructed from the textbook Math Algebra 2 by Prentice Hall which is used as one of the three full-year subject-specific courses \cite{chavez2015third}. We construct the knowledge graph by treating each unit in a chapter as a concept. To establish the relationship such as prerequisite between concepts, we use the "GO for Help" indicators provided by the authors. This indicator appears at the beginning of each unit and signals where students can get help if they are unable to understand the current unit. In Figure \ref{fig:kg-math} we show an overview of the knowledge graph constructed for units 1 to 4 for chapters 1 to 3. This graph provides a visual representation of the knowledge structure contained in these units.

\subsection{Question Retrieval and Ranking}
In the knowledge graphs used in this paper, advanced concepts are closer to the root node while preliminary concepts are closer to or at the leaf node. This means that leaf nodes are concepts that are basic and do not have any prerequisites (no incoming arrows). For example, in Figure \ref{fig:kg-math}, \textit{1-2 Algebraic Expressions} node can be considered as basic with no further prerequisite. This assumption has been made for simplicity of analysis. For each chosen node, a tree traversal algorithm is used to retrieve all directly connected descendant nodes, effectively identifying prerequisite knowledge areas or concepts and questions. These descendant nodes are then used as a reference pool to determine which prerequisite concepts should be considered when diagnosing student difficulties.

Each concept in the knowledge graph contain questions that evaluate the student's knowledge on that concept. To understand the variability in student needs on different kinds of question, we categorize questions into different difficulty level: easy (A), moderate (B) and hard (C). We associate easy questions with basic concepts. Moderately difficult questions are found on non-basic and non-advanced concepts. Hard questions are questions listed on advanced concepts. In this experiment, we take questions and their solutions from the textbook for concepts \textit{1-2 Algebraic Expressions}, \textit{1-3 Solving Equations} and \textit{3-2 Solving Systems Algebraically} and consider them as easy, moderate and hard questions respectively.

\subsection{Student's Impasse}
Most LLM-based applications often provide a query interface for users to type in their prompts. For this research, we assume students will interact with LLMs in the same manner: each student will input prompts describing their specific challenge or impasse on a given problem or a system can be designed to follow the student's solution and detect it \cite{nakamoto2021identifying}. However, due to lack of real students in this preliminary study, we asked experts to review the standard solution to each question and estimate the likely impasse for different types of students who attempt to solve the questions selected for each difficulty level. First, we define 3 different types of students as follows:

\begin{itemize}
    \item \textbf{Type S1:} Student gets stuck because they lack understanding in foundational topics. So, their issues trace back to more basic concepts.
    \item \textbf{Type S2:} Student has a middling understanding. They might know the prerequisites but can still find advanced topics challenging. Their issues may trace back to intermediate concepts.
    \item \textbf{Type S3:} Student is fairly advanced. They understand all the foundational concepts but occasionally find some advanced topics challenging. Their issues might not necessarily trace back to any prerequisite.
\end{itemize}

Also, we asked the experts to suggest the likely reason for the impasse from the given list of prerequisites retrieved from the concept map in Figure \ref{fig:kg-math}. The idea here is to link the impasse to the prerequisite knowledge as not all prerequisites are equally relevant to a student's specific problem.
By ranking them in order of relevance, the system can prioritize feedback and interventions, addressing the most pressing needs first and optimizing the student's learning trajectory. It is important to reiterate that the focus of this work is not to guess or estimate a student's impasse. Our main task is to evaluate to what degree can LLMs provide personalized feedback to different types of students with varying needs given the student's current knowledge state.

\subsection{LLM-based Personalized Solution Generation}
Traditional feedback systems often provide generic answers that may not address specific student needs and impasse. In this paper, we propose the use of LLM to generate personalized answers to given problems with emphasis on the impasse and traced prerequisites. This is done using the prompt \textbf{P1} below. The difference between our proposed method and direct prompt to ChatGPT is that we factor in the student's current knowledge state as traced from the knowledge graph and estimated by experts. Given this knowledge state, it is expected that the LLM generates a tailored feedback or answer to address the student's specific challenges. Each student's learning journey is unique; hence, feedback should be as well.

\begin{displayquote}
\textbf{P1:} \textit{Solve this question: \{question\}. The correct and standard solution is \{answer\}. Your solution should include detailed explanation to help this impasse:: \{impasse\}. This impasse exists because: \{ranked\_prerequisites\}?}
\end{displayquote}

\section{Experiment}
\begin{figure*}[h!]
  \centering
  \includegraphics[width=0.7\linewidth]{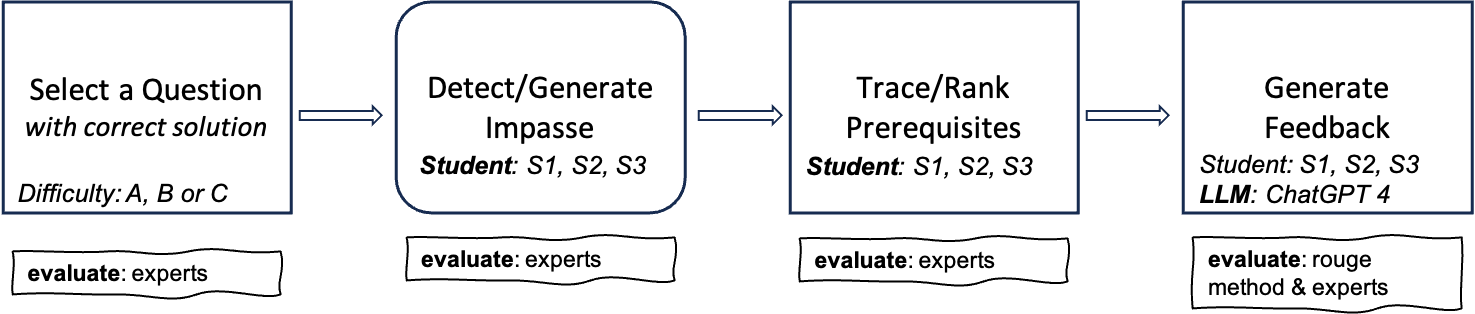}
  \caption{Experiment setup.}
  \label{fig:experiment}
\end{figure*}

We conduct a preliminary evaluation of our proposed method by setting up an experiment as shown in Figure \ref{fig:experiment}. First, we select one question from each difficulty levels Easy (A), Moderate (B) and Hard (C). For easy questions usually taken from concepts on the leaf node, we expect different types of students (S1, S2, and S3), to have similar impasse. Thus, it is acceptable (and expected) for the adapted guidance received to be similar. For moderate questions which are taken from non-leaf nodes,  the different types of student might occasionally have overlapping impasse and hence similar adapted guidance is permissible. Hard questions are taken from concepts further up the tree and closer to the root node. Given the advance level of hard questions, different student types are expected to have different learning impasse and thus, receive different feedback.

In the second stage, we generate the personalized feedback from ChatGPT4 using prompt \textit{P1}. ChatGPT was prompted once per question with the temperature set to 0.2 to ensure consistency in output. This approach is based on established practices in natural language processing (NLP), where lower temperature settings reduce randomness and increase consistency in generated responses\cite{jin2024robotgpt}. To measure the quality and relevance of the results to each student's needs, we employ the ROUGE method of evaluating text summarization. First, we compare the standard solution (S1) to each of the personalized feedback generated by ChatGPT4 for S1, S2 and S3. Next, we compare the personalized feedback to one another to test variability and level of personalization for each student. To validate the results obtained from the ROUGE method, we ask experts to evaluate them using a 5-point likert scale which will be described in the next section.

\subsection{Evaluation metrics}

We evaluate the various texts generated by ChatGPT-4 using two main methods. The first measure is the ROUGE method. The ROUGE method measures the quality of a summary by evaluating overlapping units such as n-grams, word sequences, and word pairs between a reference and candidate text. In this study, we consider the text generated by ChatGPT-4 as summaries and perform pairwise comparisons of n-grams across the different responses generated for each student type to obtain ROUGE-N and $F_1$ scores.

The use of ROUGE in this context serves to assess the degree of personalization in the feedback. Our assumption is that effective personalization in feedback should result in explanations that contain essential elements of the standard solution while varying in other aspects to address individual student needs. To provide context, we analyze the personalized feedback for different student profiles and question difficulty levels. For more challenging problems, where students might encounter various sticking points, we expect lower ROUGE scores because personalized feedback should diverge more from the standard solution to address specific areas where students struggle. A high ROUGE score would indicate a lack of personalization, as it would suggest that the feedback is too similar to the standard solution. By comparing the personalized feedback generated by ChatGPT-4 for student type S1 to that of both S2 and S3, and S2 to S3, we aim to evaluate the variability and uniqueness of the feedback provided. Similarly, we repeat the same ROUGE-N measure for the standard solution (S) versus the adapted guidance generated by ChatGPT-4 for all three student types across different question difficulty levels.

The second method of evaluation is the use of expert evaluators. 3 experts evaluated the adapted feedback generated by ChatGPT4. This expert evaluation were based on the following criteria:
\begin{itemize}
    \item \textbf{Correctness:} With respect to the question given, what is the correctness of the answer generated by ChatGPT4? It is important to check the correctness of the response generated by ChatGPT4 because prior works have reported the tendencies of LLMs to generate wrong answers \cite{wu2023autogen}. This is even more critical to ensure correctness to avoid teaching students wrong answers. The options for evaluating correctness are: 1 - Very Incorrect, 2 - Incorrect, 3 - Neither Correct nor Incorrect, 4 - Correct, 5 - Very Correct.
    \item \textbf{Precision - Addresses Student's Need:} With respect to the question given and the student's impasse, how precise is the answer generated by ChatGPT4? One of the key essence of precision education \cite{wu2021analytics} is to provide personalized guidance to students based on the best available evidence. We evaluation precision of generated feedback on the following scale: 1 - Very Imprecise, 2 - Imprecise, 3 - Neither Precise nor Imprecise, 4 - Precise, 5 - Very Precise.
    \item \textbf{Hallucinations or Tendencies:} With respect to the question given and the student's impasse, how often does the answer generated by ChatGPT4 contain information irrelevant to the student and the question? There is plausible evidence that LLMs sometimes hallucinate and fabricate false information \cite{bang2023multitask}. To ensure students are not made to spend their valuable time on irrelevant feedback. We rate the tendency of ChatGPT4 to hallucinate using the following scale: 1 - Constantly, 2 - Often, 3 - Sometimes, 4 - Rarely, 5 - Never.
    \item \textbf{Variability A vs B:} With respect to the question given and the student's impasse, how different are the answers generated by ChatGPT4 for student S1 vs S2, S1 vs S3 and S2 vs S3? Similar to the measure of precision, we use the variability metric to validate the precision of the feedback and how it applies to each student. Variability is graded on the following scale: 1 - Very Low, 2 - Low, 3 - Moderate, 4 - High, 5 - Very High.
\end{itemize}

To check the inter-rater reliability, we calculate Cohen's Kappa coefficient \cite{vieira2010cohen} across the 3 expert evaluations of question types A, B and C on the above metrics.

\subsection{A pilot user study}

Finally, we provide an additional report on a pilot user study involving university students interacting with and evaluating the proposed system previously shown in Figure \ref{fig:sys_archi}. In the pilot study, an initial survey was conducted to measure students' perception of AI tools in education and feedback seeking behavior. The participants were randomly assigned one of the 3 student profiles (S1, S2 or S3) and given a math problem to solve. For each profile, a hypothetical impasse was prepared similar to the definition of each profile type. While solving the given question, each student can seek guidance from AI-sensei through 3 options: ask for further clarification on the given question, ask a specific follow-up question or prompt AI-sensei to provide a refresher on related concepts to the given question. Participants are also asked to rate the quality of the response they received from AI-sensei on a five point scale. We collected the logs and conducted a post-survey to understand the participants' experience and feedback about the proposed architecture.

\section{Result}
In this section, we present the results from the preliminary experiments on generating adaptive guidance using ChatGPT4 and knowledge graphs.

\subsection{Easy (A) question}
\subsubsection{ROUGE evaluation}
For Easy (A) type questions, the ROUGE metrics shown in Table \ref{rouge-easy-feedback-o} depict that all three student types, S1, S2, and S3, have a relatively high degree of alignment with the standard solution. The feedback for S1 edges out slightly with the highest F1-score for ROUGE-1, implying that their misunderstandings or challenges are closely mapped to the generalized feedback. S2 follows closely, whereas S3's feedback exhibits a slightly lesser alignment. The higher scores across the board compared to the Moderate (B) type indicates that there's more consistency in the type of feedback required for simpler questions.

\begin{table}[h]
\centering
\caption{ROUGE evaluation: default solution (S) vs feedback generated by ChatGPT4 for easy question}
\label{rouge-easy-feedback-o}
\begin{tabular}{|c|c|c|c|c|}
\hline
& Metric & Recall & Precision & F-score \\
\hline
\multirow{3}{*}{S vs S1} & rouge-1 & 0.81 & 0.38 & 0.51 \\
& rouge-2 & 0.60 & 0.23 & 0.33 \\
& rouge-l & 0.77 & 0.36 & 0.49 \\
\hline
\multirow{3}{*}{S vs S2} & rouge-1 & 0.79 & 0.39 & 0.52 \\
& rouge-2 & 0.64 & 0.25 & 0.36 \\
& rouge-l & 0.79 & 0.39 & 0.52 \\
\hline
\multirow{3}{*}{S vs S3} & rouge-1 & 0.77 & 0.34 & 0.47 \\
& rouge-2 & 0.55 & 0.20 & 0.29 \\
& rouge-l & 0.77 & 0.34 & 0.47 \\
\hline
\end{tabular}
\end{table}
\begin{table}[h]
\centering
\caption{ROUGE evaluation of feedback generated by ChatGPT4 for easy question}
\label{rouge-easy-feedback}
\begin{tabular}{|c|c|c|c|c|}
\hline
& Metric & Recall & Precision & F-score\\
\hline
\multirow{3}{*}{S1 vs S2} & rouge-1 & 0.54 & 0.58 & 0.56\\
& rouge-2 & 0.38 & 0.39 & 0.39\\
& rouge-l & 0.53 & 0.57 & 0.55\\
\hline
\multirow{3}{*}{S1 vs S3} & rouge-1 & 0.53 & 0.51 & 0.52\\
& rouge-2 & 0.36 & 0.34 & 0.35\\
& rouge-l & 0.52 & 0.50 & 0.51\\
\hline
\multirow{3}{*}{S2 vs S3} & rouge-1 & 0.62 & 0.56 & 0.59\\
& rouge-2 & 0.45 & 0.42 & 0.43\\
& rouge-l & 0.61 & 0.55 & 0.58\\
\hline
\end{tabular}
\end{table}

Looking into the inter-student feedback similarities in Table \ref{rouge-easy-feedback}, the S2 and S3 pairing manifests the most overlap, with the highest F1-scores across all ROUGE metrics. This suggests a shared commonality in their challenges or misconceptions for Easy (A) type questions. The S1 and S2 pairing also demonstrates a high degree of similarity, marginally outperforming the S1 and S3 combination. These scores highlight the notion that for simpler questions, students often share more universal mistakes or areas of confusion, leading to more homogeneous feedback requirements.

\subsubsection{Expert evaluation}
From the results in Table \ref{table:expert} of the evaluation by 3 experts of the output from ChatGPT4 on easy questions for student types S1, S2, and S3, we note the following findings. The evaluations were consistent across the three student types in terms of mean scores. The three student types—S1, S2, and S3—each had a mean score of 5.00 in the metrics of "Correctness" and "Hallucination" denoting an agreement among the 3 experts that ChatGPT4 outputs were correct and never contained any hallucinations. While, all 3 experts agreed that personalized feedback on an easy question for student type S2 and S3 were precise there was slight disagreement for student type S1 on the same question. There was also slight disagreement on the overall variability of the feedback provided for each student type. 2 of the experts rated variability of feedback to be high for all 3 while one rated them to be low.

\begin{table*}[b!]
\centering
\begin{tabular}{|l|l|ccc|ccc|ccc|ccc|}
\hline
& & \multicolumn{3}{c|}{\textbf{Correctness}} & \multicolumn{3}{c|}{\textbf{Precision}} & \multicolumn{3}{c|}{\textbf{Hallucination}} & \multicolumn{3}{c|}{\textbf{Overall Variability}} \\
\cline{3-14}
\textbf{Question Type} & \textbf{Measure} & \textbf{S1} & \textbf{S2} & \textbf{S3} & \textbf{S1} & \textbf{S2} & \textbf{S3} & \textbf{S1} & \textbf{S2} & \textbf{S3} & \textbf{S1} & \textbf{S2} & \textbf{S3} \\
\hline
\multirow{2}{*}{A} & Mean & 5.00 & 5.00 & 5.00 & 4.67 & 3.00 & 5.00 & 5.00 & 5.00 & 5.00 & 3.33 & 3.33 & 3.33 \\
\cline{2-14}
& SD & 0.00 & 0.00 & 0.00 & 0.58 & 1.73 & 0.00 & 0.00 & 0.00 & 0.00 & 1.15 & 1.15 & 1.15 \\
\hline
\multirow{2}{*}{B} & Mean & 5.00 & 5.00 & 5.00 & 4.67 & 4.00 & 3.67 & 5.00 & 5.00 & 5.00 & 2.67 & 2.67 & 2.67 \\
\cline{2-14}
& SD & 0.00 & 0.00 & 0.00 & 0.58 & 1.00 & 1.15 & 0.00 & 0.00 & 0.00 & 1.15 & 1.15 & 1.15 \\
\hline
\multirow{2}{*}{C} & Mean & 4.67 & 4.67 & 5.00 & 4.00 & 4.33 & 4.67 & 4.67 & 5.00 & 5.00 & 4.00 & 4.00 & 4.00 \\
\cline{2-14}
& SD & 0.58 & 0.58 & 0.00 & 0.00 & 1.15 & 0.58 & 0.58 & 0.00 & 0.00 & 1.00 & 1.00 & 1.00 \\
\hline
\end{tabular}
\caption{Descriptive Statistics for Correctness, Precision, Hallucination, and Overall Variability Metrics}
\label{table:expert}
\end{table*}

\subsection{Moderately difficult (B) question}
\subsubsection{ROUGE evaluation}
For Moderate (B) type questions, the provided ROUGE metrics in Table \ref{rouge-moderate-feedback-o} reveal that the personalized feedback for student type S3 is most aligned with the standard solution, with the highest F1-scores across all ROUGE metrics. This indicates that the challenges and difficulties faced by S3 students for moderate questions closely align with the generalized or expected path. Feedback for student type S2 also demonstrates reasonable alignment, though not as high as that for S3. S1's feedback, conversely, has the least alignment with the standard solution, suggesting that S1 students' impasses for these questions could be more specialized or distinct from the norm.

\begin{table}[h]
\centering
\centering
\caption{ROUGE evaluation: default solution (S) vs feedback generated by ChatGPT4 for moderate question}
\label{rouge-moderate-feedback-o}
\begin{tabular}{|c|c|c|c|c|}
\hline
& Metric & Recall & Precision & F-score \\
\hline
\multirow{3}{*}{S vs S1} & rouge-1 & 0.43 & 0.20 & 0.27 \\
& rouge-2 & 0.21 & 0.09 & 0.12 \\
& rouge-l & 0.40 & 0.18 & 0.25 \\
\hline
\multirow{3}{*}{S vs S2} & rouge-1 & 0.56 & 0.27 & 0.36 \\
& rouge-2 & 0.35 & 0.16 & 0.22 \\
& rouge-l & 0.54 & 0.26 & 0.35 \\
\hline
\multirow{3}{*}{S vs S3} & rouge-1 & 0.68 & 0.28 & 0.40 \\
& rouge-2 & 0.40 & 0.14 & 0.21 \\
& rouge-l & 0.62 & 0.26 & 0.36 \\
\hline
\end{tabular}
\end{table}
\begin{table}[h]
\centering
\caption{ROUGE evaluation of feedback generated by ChatGPT4 for moderate question}
\label{rouge-moderate-feedback}
\begin{tabular}{|c|c|c|c|c|}
\hline
& Metric & Recall & Precision & F-score \\
\hline
\multirow{3}{*}{S1 vs S2} & rouge-1 & 0.46 & 0.48 & 0.47 \\
& rouge-2 & 0.20 & 0.22 & 0.21 \\
& rouge-l & 0.41 & 0.42 & 0.41 \\
\hline
\multirow{3}{*}{S1 vs S3} & rouge-1 & 0.59 & 0.54 & 0.57 \\
& rouge-2 & 0.33 & 0.29 & 0.31 \\
& rouge-l & 0.55 & 0.50 & 0.52 \\
\hline
\multirow{3}{*}{S2 vs S3} & rouge-1 & 0.48 & 0.42 & 0.45\\
& rouge-2 & 0.28 & 0.21 & 0.24\\
& rouge-l & 0.45 & 0.39 & 0.42\\
\hline
\end{tabular}
\end{table}

Considering the similarity in feedback among student types, the S1 and S3 pairing shows the most significant overlap, as reflected by the highest F1-scores across all ROUGE metrics in Table \ref{rouge-moderate-feedback}. This suggests that both S1 and S3 students might share common misunderstandings or challenges for Moderate (B) type questions. Comparatively, the S1 and S2, as well as the S2 and S3 pairings, demonstrate lesser alignment in their feedback. Among these, the S1 and S2 combination yields a slightly better similarity. Overall, while there are some overlaps in the challenges faced by different student types, the extent of similarity varies, underlining the distinct learning trajectories of these student categories.

\subsubsection{Expert evaluation}
From Table \ref{table:expert}, for moderately difficult question (Q2), all 3 experts agree that all feedback provided were very correct and that ChatGPT4 never hallucinated. However, there is mild disagreement in how precise the feedback helps the student to overcome the impasse with ratings ranging between Very Precise - 5 and Neither Imprecise or Precise - 3. In terms of the similarity of the personalized feedback, 2 experts rate the overall variability to be low across the 3 student type while 1 expert rated all 3 to be High.

\subsection{Hard (C) question}
\subsubsection{ROUGE evaluation}
For hard (B) type questions, the results highlight a variation in the personalized feedback for individual student types and the standard solution as shown in Table \ref{rouge-hard-feedback-o}. The feedback for student type S1 exhibits the highest F1-scores across all ROUGE metrics, suggesting that the guidance provided for S1 is most congruent with the standard solution. On the other hand, S3's feedback demonstrates the lowest alignment, implying that the challenges faced by this student type might be more unique or divergent from the standard path. S2's feedback falls in between, reflecting a balance between the patterns observed in S1 and S3.

\begin{table}[b!]
\centering
\caption{ROUGE evaluation: default solution (S) vs feedback generated by ChatGPT4 for hard question}
\label{rouge-hard-feedback-o}
\begin{tabular}{|c|c|c|c|c|}
\hline
& Metric & Recall & Precision & F-score \\
\hline
\multirow{3}{*}{S vs S1} & rouge-1 & 0.65 & 0.26 & 0.37 \\
& rouge-2 & 0.38 & 0.12 & 0.19 \\
& rouge-l & 0.63 & 0.25 & 0.36 \\
\hline
\multirow{3}{*}{S vs S2} & rouge-1 & 0.52 & 0.37 & 0.43 \\
& rouge-2 & 0.30 & 0.17 & 0.22 \\
& rouge-l & 0.46 & 0.33 & 0.39 \\
\hline
\multirow{3}{*}{S vs S3} & rouge-1 & 0.43 & 0.33 & 0.37 \\
& rouge-2 & 0.22 & 0.14 & 0.17 \\
& rouge-l & 0.37 & 0.29 & 0.32 \\
\hline
\end{tabular}
\end{table}
\begin{table}[h]
\centering
\caption{ROUGE evaluation of feedback generated by ChatGPT4 for hard question}
\label{rouge-hard-feedback}
\begin{tabular}{|c|c|c|c|c|}
\hline
& Metric & Recall & Precision & F-score \\
\hline
\multirow{3}{*}{S1 vs S2} & rouge-1 & 0.36 & 0.65 & 0.46 \\
& rouge-2 & 0.20 & 0.37 & 0.26 \\
& rouge-l & 0.35 & 0.64 & 0.45 \\
\hline
\multirow{3}{*}{S1 vs S3} & rouge-1 & 0.30 & 0.59 & 0.40 \\
& rouge-2 & 0.13 & 0.27 & 0.18 \\
& rouge-l & 0.29 & 0.57 & 0.39 \\
\hline
\multirow{3}{*}{S2 vs S3} & rouge-1 & 0.48 & 0.51 & 0.50 \\
& rouge-2 & 0.21 & 0.23 & 0.22 \\
& rouge-l & 0.45 & 0.49 & 0.47 \\
\hline
\end{tabular}
\end{table}

When examining the feedback overlap between student types as shown in Table \ref{rouge-hard-feedback}, the S2 and S3 pairing showcase the most significant similarity, evident from their higher F1-scores across all ROUGE metrics. This might suggest that for hard questions, S2 and S3 face more aligned challenges or misconceptions, demanding similar guidance. Contrarily, the feedback for S1 and S3, as well as for S1 and S2, shows less overlap. Of these, the S1 and S2 comparison yielded the highest F1-scores, suggesting a moderate degree of similarity in their feedback. These outcomes reaffirm the hypothesis that different student types exhibit distinct learning impasses, particularly in the context of hard questions.

\subsubsection{Expert evaluation}
For a hard question (Q3), all 3 evaluators ranked the feedback provided by ChatGPT4 as either correct (4) or very correct (5). Also, in terms of hallucination the 3 experts opined that ChatGPT4 never or rarely exhibited such traits. All 3 evaluators' rating of the precision of personalized feedback ranged between very precise and neither imprecise nor precise suggesting that in some cases, feedback may contain little personalization. This is further reinforced by the ratings received on the overall variability metric: all 3 experts rated variability to be between very high and moderate.

\subsection{General reliability of expert ratings}
We assessed the inter-rater reliability among the three evaluators using Cohen's Kappa statistic for three types of questions (A, B, and C). Cohen's Kappa is a robust measure that accounts for the agreement occurring by chance, providing a more accurate assessment of inter-rater consistency. The results indicated a Cohen's Kappa value of $0.47$ for A, $0.4$2 for B, and $0.30$ for C. According to the commonly accepted interpretation scale \cite{gray2022stakeholders}, these values correspond to moderate agreement for A and B, and fair agreement for C. The moderate agreement observed for A and B suggests a reasonable level of consistency among the raters, while the fair agreement for C highlights some variability in the ratings.

\subsection{Result from pilot user study}
A total of 6 participants took part in a pilot user study of our proposed AI sensei system.
Figure \ref{fig:pre-test-plot} shows the results from the pre-test survey. Participants' familiarity with AI and LLMs had mean scores of 3.6 and 3.4, respectively, indicating a moderate to high level of familiarity. The frequency of LLM use was relatively high, with a mean score of 4.0, suggesting frequent use. The ratings for the usefulness of LLMs in providing feedback and for asking questions varied, with mean scores of 3.0 for feedback and 3.8 for questions. The rating for using LLMs for current learning activities had a mean score of 3.0, indicating a balanced perspective on their effectiveness in this context.

\begin{figure}[h]
     \centering
     \includegraphics[width=1\linewidth]{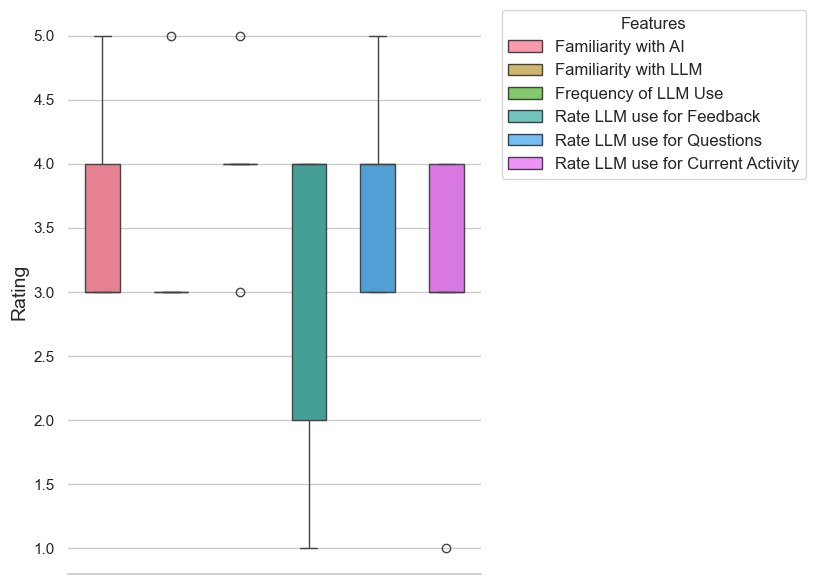}
     \caption{Participants' pre-test perspective on use of AI}
     \label{fig:pre-test-plot}
\end{figure}

The post-test survey conducted assessed the participants' perceptions of the AI sensei tool after using it as an assistant in solving math problems. Participants rated the tool on several metrics, including ease of use, correctness, usefulness, allowance for asking irrelevant questions, occurrence of AI hallucinations, and various aspects of problem-solving assistance. The mean ratings for ease of use $(\mu=3.50)$, correctness $(\mu=3.33)$, and usefulness $(\mu=3.17)$ were relatively high, indicating positive feedback from the participants. The correctness measure was also consistent with the ratings $(\mu=3.23, \sigma=1.33)$ given by the participants for each of the responses received from the AI sensei. Interestingly, participants rated AI sensei to be infrequent in hallucinations with a very low mean rating $(\mu=1.67)$ and standard deviation $(\sigma=0.52)$. Another interesting observation is that while the pre-test survey showed participants had a low perception on the use of LLM in providing feedback to learners $(\mu=3.0, \sigma=1.41)$, the post-test results showed a significant improvement in this perception $(\mu=3.67, \sigma=0.82)$. This suggests a positive effect of our proposed framework and method. This is further reinforced by some comments from the participants like:

\begin{quote}
    I also felt that AI Sensei was very helpful in helping me with "motivation issues related to study," although it did not tell me about "tomorrow's weather. In fact, it would be better to add such a choice...
\end{quote}

\begin{figure}[h!]
     \centering
     \includegraphics[width=1\linewidth]{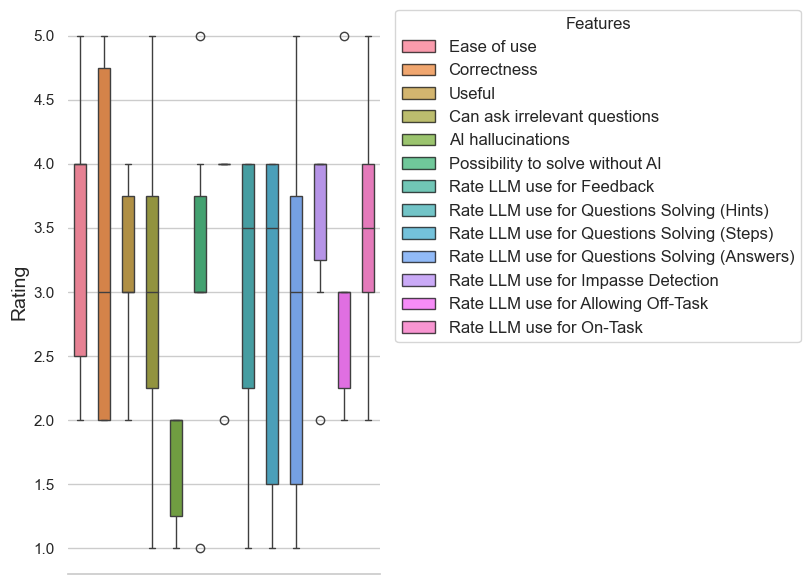}
     \caption{Participants' post-test perspective on use of AI sensei}
     \label{fig:pre-test-plot}
\end{figure}

\section{Discussions and Limitations}

For easy questions, we discovered more consistency in feedback, reflecting a relatively high degree of alignment with the standard solution. This suggests that common mistakes in simpler questions have more uniform feedback requirements. However, this pattern faded with increased complexity from moderate to hard, indicating the nuanced understanding that might be needed when addressing more challenging student queries.

A critical observation from our study is that LLMs, while advanced, are not infallible. They sometimes misinterpret context or nuances, generating feedback that might be ambiguous or inapplicable. This underlines the indispensable need for human oversight, especially in educational scenarios. Before deploying such feedback mechanisms in live classroom settings, it's crucial to integrate a layer of human validation to filter out potentially misleading or inadequate feedback. The stakes are high in education, and LLM-generated responses, if unchecked, could inadvertently foster misconceptions or hamper learning.

The ROUGE method, though robust in many text similarity applications, brings its own set of assumptions and potential errors. Its precision-recall based metrics focus on overlapping n-grams, which might not always capture the essence or context of feedback. A high ROUGE score does not unequivocally signify meaningful or contextually appropriate feedback. Similarly, a lower score does not always denote inadequacy but might point to a variation in phrasing or approach. Thus, metrics obtained should be interpreted with caution.

Results from the expert reviews also provide some valuable insights for this research. For the questions reported in our experiment, all experts rated all the answers by ChatGPT4 to be correct. However, it is important to note that when we did not provide ChatGPT with the correct standard solution, the answers or feedback generated were wrong. Also, when we asked ChatGPT4 to estimate typical impasse students would face on a given problem, it performed poorly as the impasses generated were either too generic or similar for different student profiles. Thus, we asked experts to provide possible impasses for the different student profiles based on the needs of each question. This lack of ability of LLMs like ChatGPT4 to estimate learning difficulty makes a strong case for delegating such tasks to existing pedagogical research and tools for learning analytics. We discovered that despite LLMs inability to estimate learning difficulties, they provide a decent feedback most times to address these impasses.

Jeon and Lee \cite{jeon2023large} and Kasneci et al. \cite{kasneci2023chatgpt} found that LLMs, despite their advanced capabilities, still require significant human oversight to ensure the accuracy and applicability of their feedback. Their study emphasizes the complementary role of human educators in validating and contextualizing AI-generated content to prevent the dissemination of misconceptions. This complements our recommendation for integrating human validation layers in the deployment of LLMs in educational settings.

Our study comes with its fair share of limitations. We have exclusively considered mathematics, delving into only a fraction of the vast knowledge graph on Algebra 2 for High School students. A holistic understanding would require branching out to more topics, other subjects and assessing LLM feedback across a broader academic spectrum. The study's linguistic scope was also confined to English, not considering the multitude of global students who learn in other languages. Expanding to multiple languages would offer richer insights into the universality of LLM efficacy. Also, the absence of real students in our experiment led to not capturing the true diversity and unpredictability of student impasse(s). Real-world classroom scenarios might present challenges and nuances not addressed by our simulation.

\section{Conclusion}
This paper proposed a new approach to generate personalized feedback for students by augmenting prompts to LLMs with student's knowledge state retrieved from a relevant knowledge graph. The experiments conducted simulated impasses of 3 different types of students across questions from 3 difficulty levels. The results of the feedback generated by ChatGPT4 revealed that for easy questions, all 3 types of student receive similar guidance. As the difficulty level increases, the personalization of feedback also tends to increase. However, some of the feedback generated contain some errors and sometimes not adequately catering to the student's impasse. This calls for careful supervision when using LLMs to support teaching and learning. While LLMs like ChatGPT4 showcase promising capabilities in assisting educators and personalizing feedback, they are not ready to replace human intervention. Their application, while promising, must be approached with caution, thorough oversight, and continuous validation to harness their potential effectively and responsibly in educational settings.

\section*{Acknowledgment}
This work was supported in part by the following grants: Japan Society for the Promotion of Science (JSPS) Grant-in-Aid for Fellows JP22KJ1914, (B) JP20H01722, JP23H01001, (Exploratory) JP21K19824, (B) JP22H03902 and New Energy and Industrial Technology Development Organization (NEDO) under Grant JPNP20006 and Grant JPNP18013.

\bibliographystyle{IEEEtran}
\bibliography{bare_jrnl_new_sample4}

\newpage

\begin{IEEEbiography}[{\includegraphics[width=1in,height=1.25in,clip,keepaspectratio]{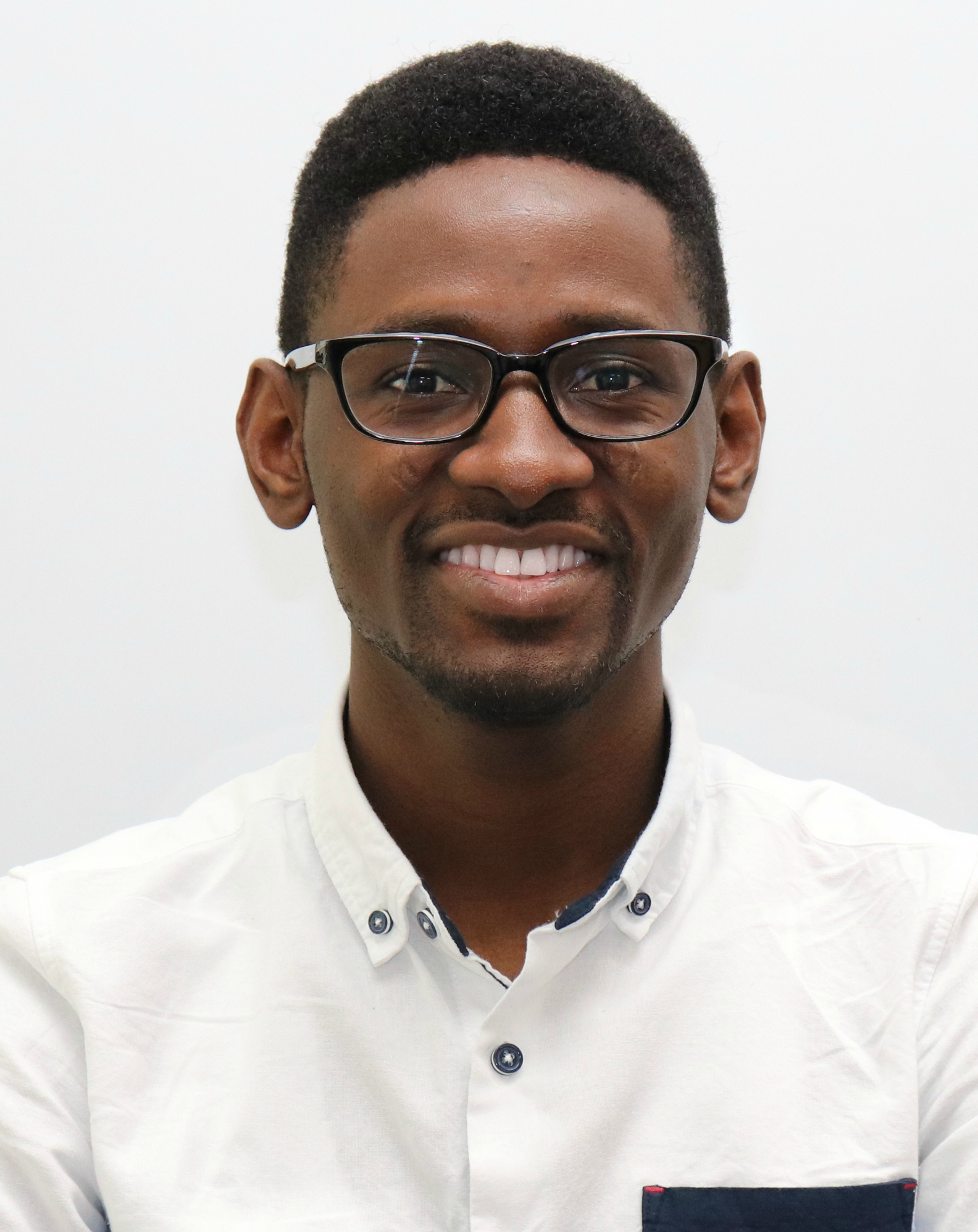}}]{Patrick Ocheja} received the BEng. degree in electronic engineering from the University of Nigeria, Nsukka, Nigeria in 2014 and the M.S. degree in Informatics from Kyoto University, Kyoto, Japan, in 2019. Dr. Patrick received his Ph.D. in Informatics from Kyoto University in 2022 and he is currently a Japan Society for the Promotion of Science (JSPS) Postdoctoral Fellow at Kyoto University, Kyoto, Japan.

Outside academia, from 2014 to 2016, he worked as a software engineer at Gidi Mobile Limited: a learning and education company in Nigeria. His research interest includes learning analytics, personalization, lifelong learning, privacy, distributed, and decentralized systems.
\end{IEEEbiography}
\vspace{-10mm}

\begin{IEEEbiography}[{\includegraphics[width=1in,height=1.25in,clip,keepaspectratio]{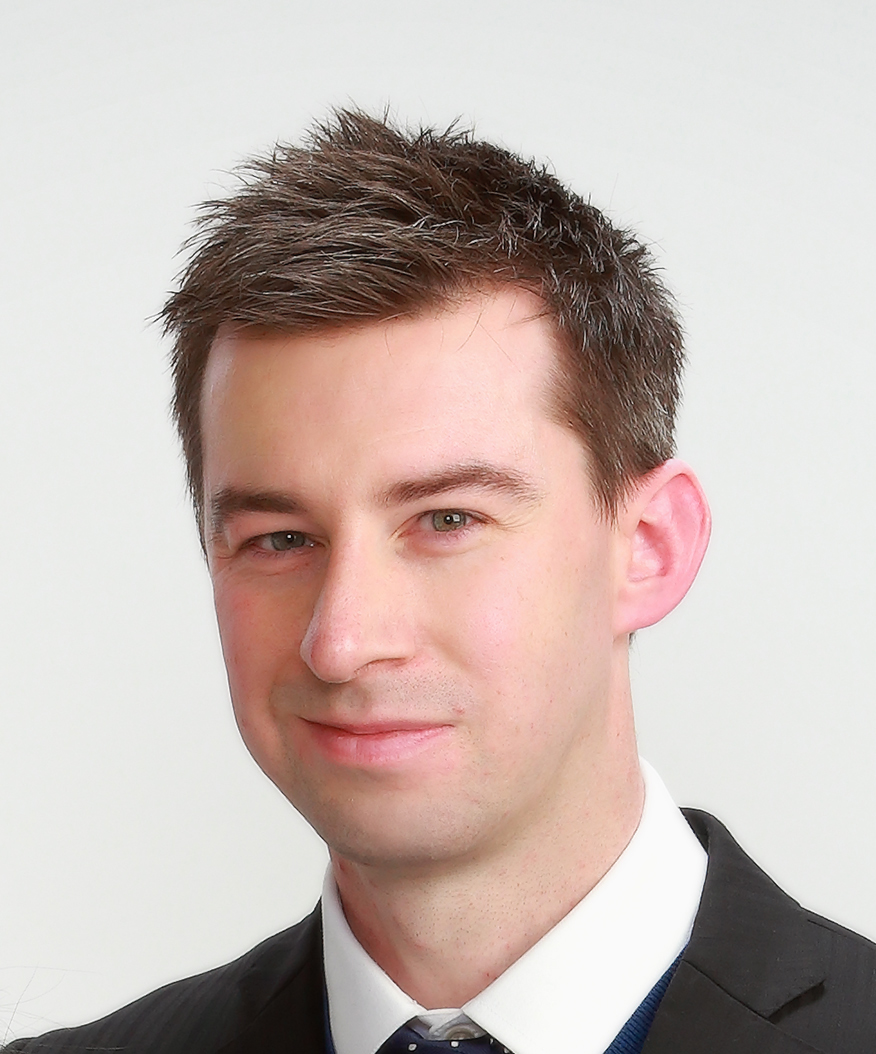}}]{Brendan Flanagan} received a Bachelor's degree from RMIT University, Master's and PhD degrees from the Graduate School of Information Science and Electrical Engineering, Kyushu University. He is currently an Associate Professor at the Academic Center for Computing and Media Studies, Kyoto University. Research interests include: learning analytics, text mining, machine learning, and language learning.
\end{IEEEbiography}
\vspace{100mm}

\newpage

\begin{IEEEbiographynophoto}{Yiling Dai} is a Program-Specific Researcher at the Academic Center for Computing and Media Studies in Kyoto University. She received her Ph.D in Informatics in Kyoto University. Her research interests include: Information retrieval, knowledge representation, recommender system, educational data mining, and learning analytics.
\end{IEEEbiographynophoto}
\vspace{20mm}

\begin{IEEEbiography}[{\includegraphics[width=1in,height=1.25in,clip,keepaspectratio]{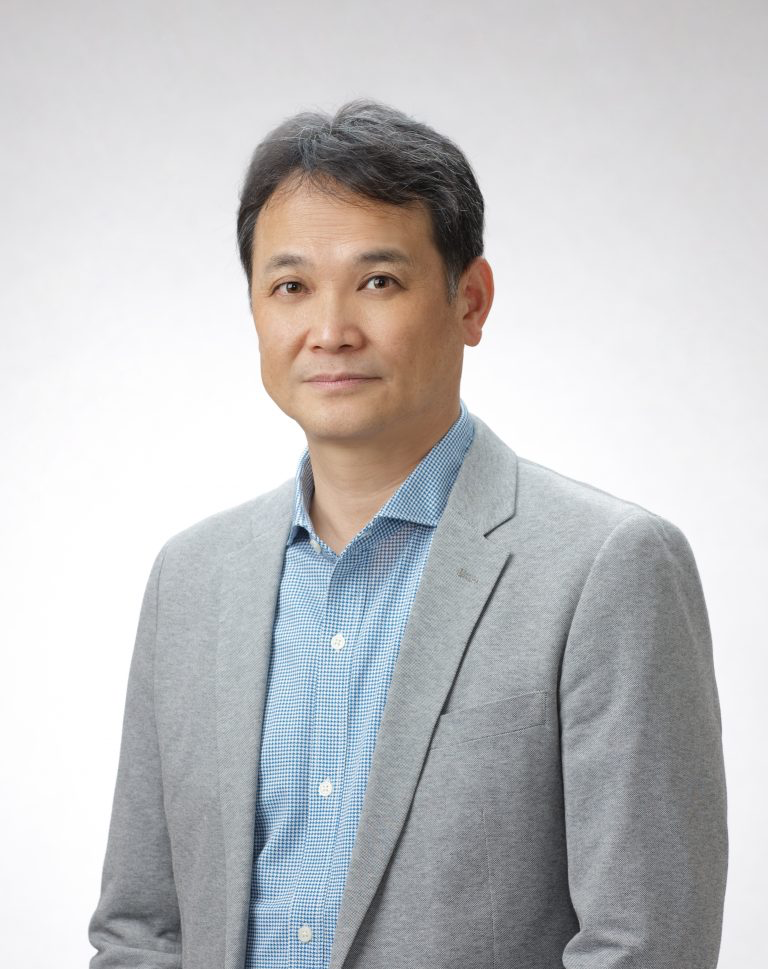}}]{Hiroaki Ogata} is a Professor at Learning and Educational Technologies Research Unit, the Academic Center for Computing and Media Studies, and the Graduate School of Informatics at Kyoto University, Japan.

His research interests include Learning Analytics, Evidence-Based Education, Educational data mining, Educational Data Science, Computer Supported Ubiquitous and Mobile Learning, CSCL (Computer Supported Collaborative Learning), CSCW (Computer Supported Collaborative Writing), CALL (Computer Assisted Language Learning), CSSN (Computer Supported Social Networking), Knowledge Awareness, Personalized, adaptive  and smart learning environments. Currently, Professor Ogata is leading a research project on the development of infrastructure for learning analytics and educational data science.
\end{IEEEbiography}
\vspace{-10mm}
\enlargethispage{-9.5cm}
\vfill

\end{document}